# Inteligencia artificial para la multi-clasificación de fauna en fotografías automáticas utilizadas en investigación científica

Gonzalez Federico[1], Viera Leonel[1], Soler Rosina[2], Chiarvetto Peralta Lucila[1], Gel Matías[1], Bustamante Gimena[2], Montaldo Abril[1], Rigoni Brian[1], Perez Ignacio[1]

[1] Instituto de Desarrollo Económico e Innovación
Universidad Nacional de Tierra del Fuego, Antártida e Islas del Atlántico Sur
IDEI - UNTDF
Ushuaia, Tierra del Fuego

[2] Centro Austral de Investigaciones Científicas
CADIC - CONICET
Ushuaia, Tierra del Fuego

{fgonzalez, lviera, mgel, lchiarvetto}@untdf.edu.ar
{rosina.soler, gimenabustamante}@conicet.gov.ar
{abril.4545, brian.rigoni1, ignacioperez583}@gmail.com

## RESUMEN

El manejo de ambientes naturales, ya sea para conservación o producción, requiere de una profunda comprensión de la vida silvestre. El número, la ubicación y el comportamiento de los animales salvajes es uno de los principales objetos de estudio en ecología y vida silvestre. El uso de cámaras trampa ofrece la oportunidad de recopilar rápidamente grandes cantidades de fotografías que -sin la presencia humana- registran a la fauna en su hábitat natural, evitando factores que alteren su comportamiento.

En Tierra del Fuego, Argentina, se desarrollan investigaciones sobre el uso del bosque por parte de distintos herbívoros (guanacos, vacas, ovejas) para optimizar el manejo y proteger dichos ecosistemas naturales.

Si bien las cámaras trampa permiten la obtención de millones de imágenes, la interpretación de tales fotografías representa un problema de escala para el procesamiento manual. Así, gran parte del valioso conocimiento en estos enormes repositorios de datos sigue sin ser aprovechado.

Las Redes Neuronales y el Deep Learning son áreas de estudio dentro la Inteligencia Artificial, durante la última década estas dos disciplinas han hecho cuantiosos aportes en el ámbito del reconocimiento de imágenes de gran relevancia a nivel mundial.

Los estudios ecológicos y de conservación de la vida silvestre, pueden combinarse con estas nuevas tecnologías para extraer información importante a partir de las fotografías obtenidas por cámaras trampa, con el objeto de aportar a la comprensión de distintos procesos naturales y mejorar el manejo de las áreas silvestres implicadas.

Nuestro proyecto busca desarrollar modelos de redes neuronales para clasificar especies de animales en fotografías obtenidas mediante cámaras trampa, para resolver problemas de gran volumen en investigación científica.

**Palabras clave:** deep learning ; computer vision ; trap camera ; vida silvestre

## CONTEXTO

Este proyecto forma parte de la línea de investigación "Innovación en sistemas de software" del área 8 "Desarrollo Informático", del Instituto de Desarrollo Económico e Innovación de la Universidad Nacional de Tierra del Fuego (IDEI-UNTDF).

Nuestro grupo de investigación viene trabajando temas afines desde 2017 y en 2020 comenzamos con la interpretación de cámaras trampa en el proyecto interno del IDEI "Inteligencia Artificial para la identificación de fauna en fotografías automáticas utilizadas en investigación científica". Recientemente, hemos postulado a un proyecto PID-UNTDF, que esperamos sea aprobado próximamente y su ejecución se dará a lo largo de 2022 y hasta mediados de 2023.

## INTRODUCCIÓN

El manejo de ambientes naturales, ya sea para conservación o producción, requiere el conocimiento del comportamiento de la vida silvestre. El número, la ubicación y el comportamiento de los animales silvestres es uno de los principales objetos de estudio en ecología y manejo de la vida silvestre (Silveira et al. 2003, Fegraus et al. 2011, Palmer y Packer 2018). El uso de cámaras con sensores de movimiento en hábitats naturales -llamadas cámaras trampa- ha transformado la investigación en ecología y conservación de la vida silvestre en las últimas dos décadas (O'Connell et al. 2010). Estas se han convertido en una herramienta esencial para los ecologistas, permitiéndoles estudiar el tamaño y la distribución de las poblaciones (Silveira et al. 2003) y evaluar el uso del hábitat. Si bien permiten tomar millones de imágenes (Fegraus et al. 2011), la clasificación y extracción de información (datos) es tradicionalmente realizada por humanos (es decir, expertos o una comunidad de voluntarios), es lenta y costosa debido a su procesamiento manual. Así, gran parte del valioso conocimiento en estos grandes repositorios de datos sigue sin ser aprovechado.

Actualmente, estas cámaras tienen sensores de movimiento que se activan con la presencia de un animal, registrando una imagen digital por cada situación detectada. Dichos sensores pueden configurarse para ser más o menos sensibles. Técnicamente, el sensor de movimiento reacciona a los cambios de luces, una configuración sensible produce una mayor cantidad de falsos positivos, una configuración menos sensible es propensa a no fotografiar todas las situaciones necesarias.

Para evitar el sub-muestreo de individuos, es común configurar la cámara en un rango de mayor sensibilidad; esto lleva a obtener una inmensa cantidad de registros innecesarios, ya que cualquier movimiento será interpretado como "necesario de fotografiar". Así, por ejemplo, dentro de un bosque los movimientos de ramas u otra vegetación activan el sensor y la cámara acciona el diafragma, pero probablemente ningún animal haya estado presente. Lo mismo ocurre cuando una nube tapa el sol repentinamente, cuando nieva, etc.

Otro de los problemas típicos es la presencia de animales que no son de interés para el estudio (ej, aves), que inevitablemente se cruzan delante del lente y accionan la cámara. Finalmente, muchas de estas cámaras tienen capacidad de hacer fotografías nocturnas gracias a tecnología de infrarrojos, que si bien producen imágenes en blanco y negro, son de gran utilidad, ya que muchos animales tienen hábitos nocturnos.

Las cámaras trampa pueden pasar meses en el campo trabajando automáticamente, cuentan con baterías de larga duración y con memorias digitales de gran capacidad. Dependiendo del ambiente donde se instale el equipo, de las condiciones climáticas, el tipo de animales esperados y la sensibilidad configurada, es común que una de cada mil fotografías tomadas sea de interés para el estudio en cuestión. Si a esto se le suma la cantidad de tiempo que la cámara pase activa en el campo (semanas, meses) y la cantidad de cámaras trabajando en paralelo dentro de una misma investigación, el procesamiento de las imágenes y la tarea de encontrar manualmente las fotografías útiles para el estudio, puede convertirse en un trabajo abrumador o incluso imposible.

La Inteligencia Artificial (IA) ha tenido un importante crecimiento y popularidad durante la última década, y particularmente durante los últimos cinco años, gracias a un innovador enfoque que dio impulso al Deep Learning y las Redes Neuronales (Goodfellow et al. 2016, He et al. 2016), junto con una importante capacidad de procesamiento gracias a la invención de las GPU (Graphics Processing Unit).

Los modelos basados en redes neuronales con aprendizaje profundo son capaces de identificar patrones en imágenes, obtener conclusiones basadas en un denominador común y generar un modelo que representa a cierto grupo de imágenes con un patrón similar (Simonyan y Zisserman 2014). Luego, dada una imagen cualquiera, dicho modelo puede utilizarse para reconocer si esa imagen coincide en su representación con su patrón definido.

Los estudios ecológicos y de conservación de la vida silvestre, pueden combinarse con la IA para extraer información importante a partir de las fotografías obtenidas por cámaras trampa, con el objeto de aportar a la comprensión de distintos procesos naturales y mejorar el manejo de las áreas silvestres implicadas.

Actualmente, se encuentra en desarrollo el proyecto "Evaluación del impacto diferencial de herbívoros nativos y domésticos en bosques de *Nothofagus antarctica* (ñire) con producción ganadera en Tierra del Fuego" a cargo de la Dra. Rosina Soler, investigadora del Laboratorio de Recursos Agroforestales del CADIC-CONICET en Ushuaia. El proyecto se desarrolla en bosques nativos del centro de la provincia, en cercanías del Municipio de Tolhuin. El estudio analiza la interacción planta-herbívoro en los bosques de ñire, a partir de la respuesta de la regeneración arbórea (densidad, crecimiento, supervivencia) a la exclusión del ganado y de los herbívoros nativos. El proyecto ha recolectado más de 150.000 fotos en el período 2015-2017 utilizando un set de 24 cámaras trampa.

Esta cantidad de información representa un alto costo de análisis manual de imágenes, ralentiza el trabajo y dificulta la obtención de resultados. Durante 2020 y parte de 2021 se desarrolló también el proyecto "Inteligencia Artificial para la identificación de fauna en fotografías automáticas utilizadas en investigación científica", bajo la dirección de Federico Gonzalez y con prácticamente el mismo grupo de trabajo de este proyecto. Con la ayuda del Deep Learning y la IA, el proyecto permitió separar fotos "con animales" y "sin animales" de la mencionada base de datos.

La nueva etapa de estudio que abordaremos próximamente, pretende conocer más detalles sobre las especies encontradas en cada una de las 150.000 imágenes. Ya no sólo limitarse a clasificar presencia-ausencia de animal, sino identificar qué tipo de especies hay en cada imagen y cuántos individuos de cada una.

Más allá de lo positivo que este proyecto podría tener en el plano local, los resultados obtenidos de esta colaboración que se plantea entre el CADIC y la UNTDF tendrían la capacidad de ser transferibles globalmente a otros estudios similares en los que se utilicen este tipo de cámaras y donde se requiera clasificar imágenes para identificar fauna silvestre.

Finalmente, también cabe destacar que en la carrera de Licenciatura en Sistemas de la UNTDF no existe ninguna materia que trate temas de IA en profundidad, por lo tanto los alumnos involucrados en este proyecto de investigación tendrán la posibilidad de obtener conocimiento de gran valor para su vida profesional.

## LÍNEAS DE INVESTIGACIÓN

Nuestro trabajo se enmarca en la Inteligencia Artificial como una gran línea de investigación, aunque nuestro foco de atención se encuentra específicamente en el Deep Learning (aprendizaje profundo) y en Computer Vision (visión por computador) como ejes específicos.

## RESULTADOS Y OBJETIVOS

Nuestro proyecto busca desarrollar modelos de redes neuronales basados en Deep Learning para clasificar animales en fotografías obtenidas mediante cámaras trampa, haciendo uso de la IAnpara resolver problemas de gran volumen en investigación científica.

En 2021 hemos podido clasificar la base de datos con 150.000 imágenes, identificando en cada una la presencia-ausencia de animales con una precisión del 85% aprox. y con una variación según los ambientes naturales, la época del año y la luz disponible (día y noche) de ± 10%.

Nuestro próximo objetivo pretende especializar el modelo anterior, para clasificar las imágenes según especie animal, con un nivel de precisión similar al que una persona entrenada podría alcanzar mediante separación manual.

Luego esperamos desarrollar un modelo de red neuronal capaz de generalizarse a otros ambientes naturales y otras especies animales. Por lo cual, los resultados obtenidos tendrían la capacidad de ser transferibles globalmente a otros estudios similares.

Desde un punto de vista académico se espera aportar al conocimiento de los docentes-investigadores que participan directamente del proyecto, quienes actualmente se encuentran realizando estudios de Doctorado y Maestría; pero también para los alumnos participantes quienes manifiestan interés en aplicarlos en sus propuestas de tesis de licenciatura.

Finalmente, se espera volcar localmente lo aprendido en un taller de presentación de resultados, que estará orientado a profesores de la UNTDF, investigadores del CADIC-CONICET y otros profesionales interesados.

## FORMACIÓN DE RECURSOS HUMANOS

El equipo de trabajo está compuesto por personas provenientes de diferentes disciplinas y antecedentes. En el campo de la educación, tres profesores adjuntos con cargo docente-investigador, un asistente y tres estudiantes: todos ellos vinculados a la carrera de Lic. en Informática en la UNTDF.

A su vez, por parte de CADIC-CONICET, participan la Dra. Rosina Soler y la Ing. Gimena Bustamante. La Dra. Rosina Soler es bióloga, investigadora adjunta de CONICET y dirige el proyecto "Evaluación del impacto diferencial de herbívoros nativos y domésticos en bosques de *Nothofagus antarctica* (ñire) con producción ganadera en Tierra del Fuego", desde donde se genera la base de datos que utilizamos en este trabajo. La Ing. Gimena Bustamente está finalizando su Doctorado en Ciencias Agrarias y Forestales en la UNLP.

Mientras que la arista informática cuenta con perfiles de diferentes especialidades. El Lic. Federico González quien dirige el proyecto, tiene un Máster en Ciudades Inteligentes por la Univ. de Girona (UdG), España y está desarrollando su Doctorado en Inteligencia Artificial en la Univ. de Barcelona (UB).

Mientras que el Lic. Leonel Viera -co-director del proyecto- tiene la expectativa de finalizar su Máster en Inteligencia de Datos y BigData en la Univ. Nacional de La Plata (UNLP) dentro del marco de nuestro proyecto.

El Lic. Matías Gel está también desarrollando su Máster en Inteligencia de Datos y BigData en la UNLP.

La Lic. Lucila Chiarvetto Peralta está finalizando su Doctorado en Ciencias de la Computación en la Univ. Nacional del Sur (UNS).

Por último, los alumnos Abril Montaldo, Ignacio Perez y Brian Rigoni son Analistas Universitarios en Sistemas y están cursando su último año de Lic. en Informática, de quienes se espera que puedan comenzar su tesina de grado en temas afines al proyecto.

## REFERENCIAS

Deng J, Dong W, et al. (2009) Imagenet: A large-scale hierarchical image database. 2009 IEEE Conference on Computer Vision and Pattern Recognition (CVPR), Miami, FL, pp. 248-255.

Fegraus EH, Lin K, et al. (2011) Data acquisition and management software for camera trap data: A case study from the team network. Ecol Inform 6(6): 345-353.

Goodfellow I, Bengio Y, Courville A (2016) Deep Learning. MIT Press, Cambridge, MA, USA.

He K, Zhang X, Ren S, Sun J (2016) Deep residual learning for image recognition. 2016 IEEE Conference on Computer Vision and Pattern Recognition (CVPR), Las Vegas, NV, 2016, pp.

Krizhevsky A, Sutskever I, Hinton GE (2012) Imagenet classification with deep convolutional neural networks. 2012 Advances in Neural Information Processing Systems (NIPS).

LeCun YA, Bottou L, Orr GB, Müller KR (2012) Efficient backprop in Neural networks: Tricks of the trade. In: Montavon G, Orr GB, Müller KR. (eds) Lecture Notes in Computer Science, vol 7700. Springer, Berlin.

Stuart J. Russell, Peter Norvig (2010) Artificial Intelligence: A Modern Approach, Third Edition, Prentice Hall ISBN 9780136042594.

O'Connell AF, Nichols JD, Karanth KU (2010) Camera Traps in Animal Ecology: Methods and Analyses. Springer, Tokyo, Japan.

Palmer MS, Packer C (2018) Giraffe bed and breakfast: Camera traps reveal Tanzanian yellow-billed oxpeckers roosting on their large mammalian hosts. Afr J Ecol 56(4): 882-884.

Silveira L, Jacomo AT, Diniz-Filho JAF (2003) Camera trap, line transect census and track surveys: A comparative evaluation. Biol Conserv 114(3): 351-355.

Simonyan K, Zisserman A (2014) Very deep convolutional networks for large-scale image recognition. arXiv preprint arXiv:1409.1556

Szegedy C, Liu W, et al. (2015) Going deeper with convolutions. 2015 IEEE Conference on Computer Vision and Pattern Recognition (CVPR), Boston, MA, 2015, pp.

Wiesler S, Ney H (2011) A convergence analysis of log-linear training. 2011 Advances in Neural Information Processing Systems (NIPS).

Yosinski J, Clune J, Bengio Y, Lipson H (2014) How transferable are features in deep neural networks? 2014 Advances in Neural Information Processing Systems (NIPS).